\documentclass{article}
\usepackage{graphicx} 
\usepackage[table]{xcolor}
\usepackage{color, colortbl}

\usepackage{authblk}
\providecommand{\keywords}[1]{\textbf{\textit{Keywords:}} #1}

\title{Causality from Bottom to Top: A Survey} 
\author[1]{Abraham Itzhak Weinberg}
\author[2]{Cristiano Premebida}
\author[3]{Diego Resende Faria}
\affil[1]{AI-WEINBERG, AI Experts, Tel Aviv, Israel, aviw2010@gmail.com}
\affil[2]{University of Coimbra, Dept of Electrical and Computer Engineering, Institute of Systems and Robotics, Coimbra, Portugal}
\affil[3]{School of Physics, Engineering and Computer Science, University of Hertfordshire, Hatfield, Hertfordshire AL10 9AB, U.K.}

\begin{document}
\maketitle
\begin{abstract}
Causality has become a fundamental approach for explaining the relationships between events, phenomena, and outcomes in various fields of study. It has invaded various fields and applications, such as medicine, healthcare, economics, finance, fraud detection, cybersecurity, education, public policy, recommender systems,  anomaly detection, robotics, control, sociology, marketing, and advertising. In this paper, we survey its development over the past five decades, shedding light on the differences between causality and other approaches, as well as the preconditions for using it. Furthermore, the paper illustrates how causality interacts with new approaches such as Artificial Intelligence (AI), Generative AI (GAI), Machine and Deep Learning, Reinforcement Learning (RL), and Fuzzy Logic. We study the impact of causality on various fields, its contribution, and its interaction with state-of-the-art approaches. Additionally, the paper exemplifies the trustworthiness and explainability of causality models. We offer several ways to evaluate causality models and discuss future directions.
\end{abstract}


\section{Introduction}
Causality is one of the fundamental ways to explain phenomena. It is used by human from the dawn of history as a way for explaining results, behaviors and other facts. 
Causality can be defined as a relationship between an event (called the cause) and a second event (called the effect), where the cause brings about the effect or directly influences its occurrence \cite{halpern2015modification}.
In addition, the intuitive nature of causality makes it a common method for young children to explain the reasons behind an effect. \\
Causality can be divided hierarchy into layers or rungs such as seeing, doing and imagining \cite{pearl2019seven,bareinboim2022pearl,ibeling2021icard}. These are related respectively to association, intervention, and counterfactuals. Each level can be differentiated by its activities and the answers it can provide to relevant question \cite{shpitser2008complete}. The activities and questions are added to the precedence level. 
There are several key characteristics of causality that make it a popular concept. It explains relationships between phenomena, providing a framework for understanding why one event or thing leads to another \cite{illari2014causality}. Looking for causes helps make sense of the world. Causality also allows for prediction and control, as understanding the cause of something enables us to potentially predict its future occurrences and manipulate causal factors \cite{jonassen2008designing}. Furthermore, causality satisfies a fundamental psychological need, as humans innately seek order, purpose, and patterns in the complex world around us \cite{illari2014causality}. It provides a clear connection between events, meeting the human need for explanation and comprehension. Moreover, it facilitates learning and decision making by enhancing knowledge and enabling better-informed choices \cite{alhadad2018visualizing}. By understanding causes, individuals can learn from past experiences to avoid negative consequences or replicate positive outcomes. Causality aligns with human common sense, as it corresponds well to everyday observations of physical and natural processes. Additionally, causality is scientifically useful as it enables scientists to systematically investigate phenomena, form testable hypotheses, and advance fundamental theories. Causality's integral role in the scientific method further underscores its significance in scientific inquiry. In summary, causality is popular because it provides structure, predictability, understanding, and a sense of control, all of which are compelling and psychologically rewarding for human minds.\\
Throughout history, the concept of causality has been explored and developed by various philosophers and thinkers. In ancient Greece, philosophers such as Plato and Aristotle distinguished between formal and material causes \cite{friedman2001dynamics}, while Hellenistic thinkers delved into the realms of chance versus determinism \cite{de2007chance}. In the Middle Ages, scholars built upon Aristotelian foundations to investigate causation in relation to philosophy, theology, and physics \cite{rubenstein2004aristotle}. The early modern period saw Francis Bacon proposing an empiricist theory of causality based on regular succession \cite{baconempiricism,mulaik1987brief}, and Rene Descartes articulating deterministic causation through laws of nature \cite{descartes2007correspondence}. \\
The 18th century brought David Hume's argument that human perception is limited to observing constant conjunction rather than necessary connection between causes and effects \cite{strawson2002david}, and Immanuel Kant introduced transcendental idealism in relation to causality \cite{allais2004kant}. The 19th century witnessed advancements in statistics and probabilistic reasoning, with John Stuart Mill developing a ``Boolean canon of causation for systematic investigation of causal relationships" \cite{caramani2008introduction}. In the early 20th century, Bertrand Russell analyzed causal propositions, and the advent of quantum mechanics challenged determinism, sparking debates \cite{von2004explanation}. Carl Hempel's deductive-nomological model of explanation and the formalization of probabilistic causation using conditional independence through Bayesian networks emerged in the late 20th century \cite{mckeown1999case}. \\
In the modern era, causality has become integral to fields such as Machine Learning (ML), economics, and statistics. New frameworks for discovery from data, including interventions, counterfactuals, and causal calculus, have been introduced.
Key developments include the formulation of causality using probability theory and graphical models, such as Bayesian networks, by researchers like Pearl \cite{pearl1987embracing} and Spirtes \cite{spirtes2000causation,glymour1987discovering}. Some additional influential researchers among others, in the field of causality include Robins and Rubin \cite{didelez2022perspective}, Neyman \cite{sekhon2008neyman}, Zhang \cite{spirtes2016causal}  Gelman \cite{gelman2011causality}, Mooji \cite{mooij2016distinguishing}, Athey, Imbens \cite{athey2015machine}, Card \cite{card1999causal}, and Angrist \cite{angrist1996identification}.\\
Throughout modern times, the field of causality has experienced significant developments, particularly in its relationship with ML. Here is a rough timeline highlighting some major milestones, as shown in Figure~\ref{fig:CausalityTimeline}:
\begin{figure}[!ht]
	\begin{center}
		\includegraphics
		[scale=0.4]{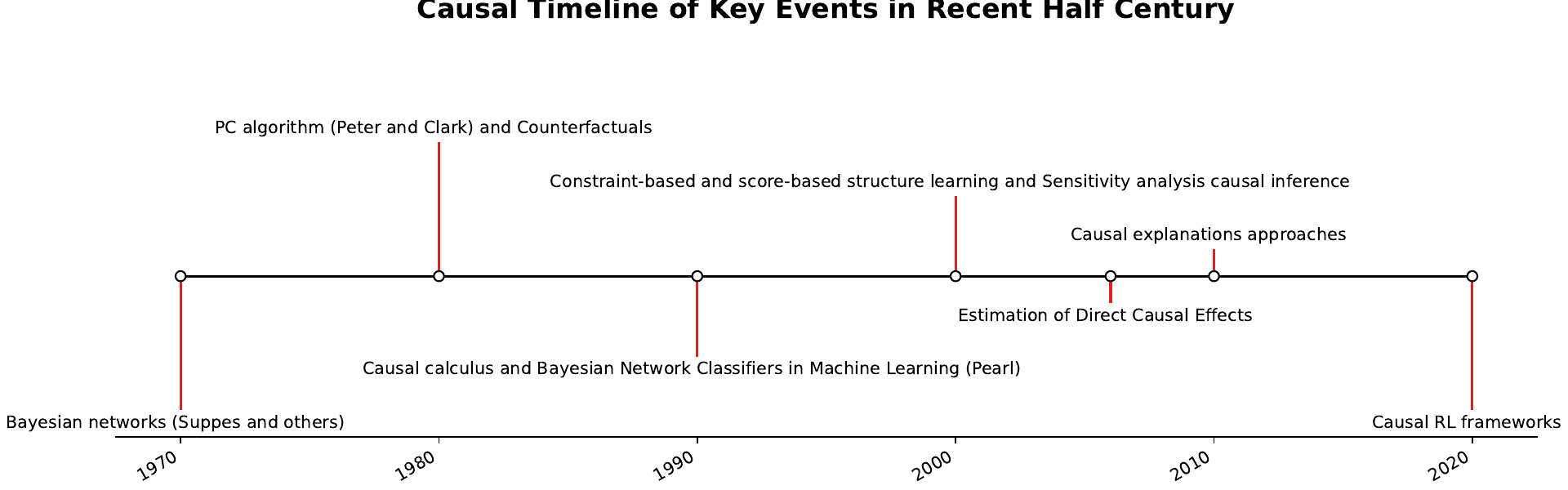}
	\end{center}
	\caption{Causality timeline with the key milestones over the last 50 years.}
	\label{fig:CausalityTimeline}
\end{figure}
In the 1970s, researchers, including Suppes and others, laid probabilistic foundations for causality by introducing Bayesian networks \cite{reiss2018suppes,suppes1974measurement}.
Moving into the 1980s, Spirtes et al. developed the PC algorithm (named after
its authors Peter and Clark) \cite{spirtes2016causal,spirtes2000causation,raghu2018comparison}, a crucial advancement for discovering causal structures. Additionally, Neyman and Rubin \cite{brady2002models,sekhon2008neyman} formalized the potential outcomes framework, which provided a solid basis for modeling causal effects using counterfactuals.
The 1990s witnessed the progress of causal calculus based on structural models and graphical criteria, pioneered by Pearl and his colleagues \cite{pearl1995causal,pearl2015trygve}. This decade also saw the application of Bayesian network classifiers in ML domains, expanding the use of causal inference methodologies \cite{cheng2013comparing,peng1987probabilistic}.\\
Advancements continued in the 2000s, with the maturation of constraint-based and score-based structure learning algorithms for causal discovery. These algorithms played a vital role in various fields, including epidemiology, where causal inference methods were widely adopted \cite{weed2000interpreting}. The notions of interventions and counterfactuals have been rigorously established, clarifying assumptions in observational studies \cite{galles1998axiomatic,vennekens2010embracing}.
An additional publication in 2000, by Robins et al. \cite{robins2000sensitivity} introduced the concept of sensitivity analysis, a technique utilized to assess the robustness of causal inferences to different assumptions or scenarios that may affect the causal relationship. In this article, we will delve into the topic and explore various common approaches and tools for conducting a sensitivity analysis in your study of causal inference.\\
In 2006, Petersen et al. \cite{petersen2006estimation} made a significant contribution by introducing a perspective on causal effect estimation in this context. Estimation of direct effects continues to be a crucial aspect of research aimed at understanding mechanistic pathways, including the ways in which exposures lead to the development or prevention of diseases, among other scenarios.
The 2010s marked a period of prolific developments in machine learning interpretability and explainability techniques. Causal explanations utilizing techniques like Shapley values \cite{heskes2020causal}, Local Interpretable Model-agnostic Explanations (LIME) \cite{schwab2019cxplain}, and anchors emerged within the ML community \cite{schaffer2019anchoring,davey2014mendelian}. In addition, game-theoretic approaches have provided causal explanations for complex ML models \cite{gurevich2012role}. It is worth noting that causal explanations aim to explain the behavior of the model, not necessarily the phenomenon that the model describes. The latter can be only concluded if the model of interest allows us to identify the causal query of interest. Moreover, explainability techniques like Shapley-values-based SHAP can lead to misleading results, regardless of whether we aim to explain the model's behavior or the mechanism behind the modeled phenomenon\cite{huang2023inadequacy}.
Causal discovery has been used in biology and genetics \cite{foraita2020causal,kelly2022,ness2017bayesian} as well as industrial applications. Moreover, causal ML found practical applications in domains such as healthcare \cite{glymour2019review} and recommendation systems \cite{gao2022causal,Goldenberg_2020,moraes2023uplift}. The development and increase of computational resources enabled  using distributed algorithms for causal discovery from massive multidimensional datasets \cite{glymour2019review}.
Causal reasoning is also being utilized to study and mitigate unfair biases and discrimination in ML models and algorithms, known as causal fairness \cite{xu2019achieving,gultchin2023casual,plecko2022causal}.
From 2020 onwards, new frontiers emerged in the field of causality. Causal RL frameworks were introduced, combining RL with causal reasoning \cite{bareinboim_forney_pearl_2015,lee2018structural,lampinen2022tell,lattimore2016causal,richens2024robust}. Causal RL has introduced causal environment models for offline evaluation, safe exploration, and transfer learning \cite{boustati2021transfer,edmonds2020theory}.
Causal fairness methods were developed to address and mitigate bias in algorithms \cite{carey2022causal,plecko2022causal,gultchin2023casual}. Researchers explored the discovery of causality from the combination of diverse datasets. Additionally, causal environment models were introduced to supplement value functions \cite{chen2022causation}.\\
Due to the COVID-19 outbreak, there can be found an increase in usage and development of algorithms for causal discovery from observational data, enabling studies in genetics and epidemiology.
Since causality is intuitive it holds the trustworthiness characteristics. This enables causality to support decision makers more than black box AI models and hence to invade more smoothly into organizations and institutions. This can be one of the reasons that usage of causality has expanded over the years and invaded into many applications and market segments. However, there is still a gap between human and AI/ML causality that will be discussed in detail in the following sections.\\  
According to Pearl \cite{pearl2009causal}, there is a distinction between causal inference and causality. Causality refers to the philosophical concept of one event or thing (the cause) being responsible for producing another event or thing (the effect), and the nature of causal relationships. 
Causal inference is the process of using statistical and computational techniques, experimental, mixed or observational data, and logical reasoning to quantify the strength of causal effects. It aims to determine causal relationships and effects between variables.
There is also a difference between causal inference and causal discovery \cite{zhang2006causal}. Causal discovery typically involves using data and statistical or computational techniques to retrieve information about causal relationships from the data, while causal inference utilizes the knowledge about causal structure to quantify the strengths of causal relationships and make predictions. Note that this terminology is not always used consistently in the literature.

\section{Causality Characteristics and Uniqueness}
\begin{figure}[!ht]
	\begin{center}
		\includegraphics
        [trim=3.5cm 7.5cm 1.0cm 3.5cm, scale=0.5]{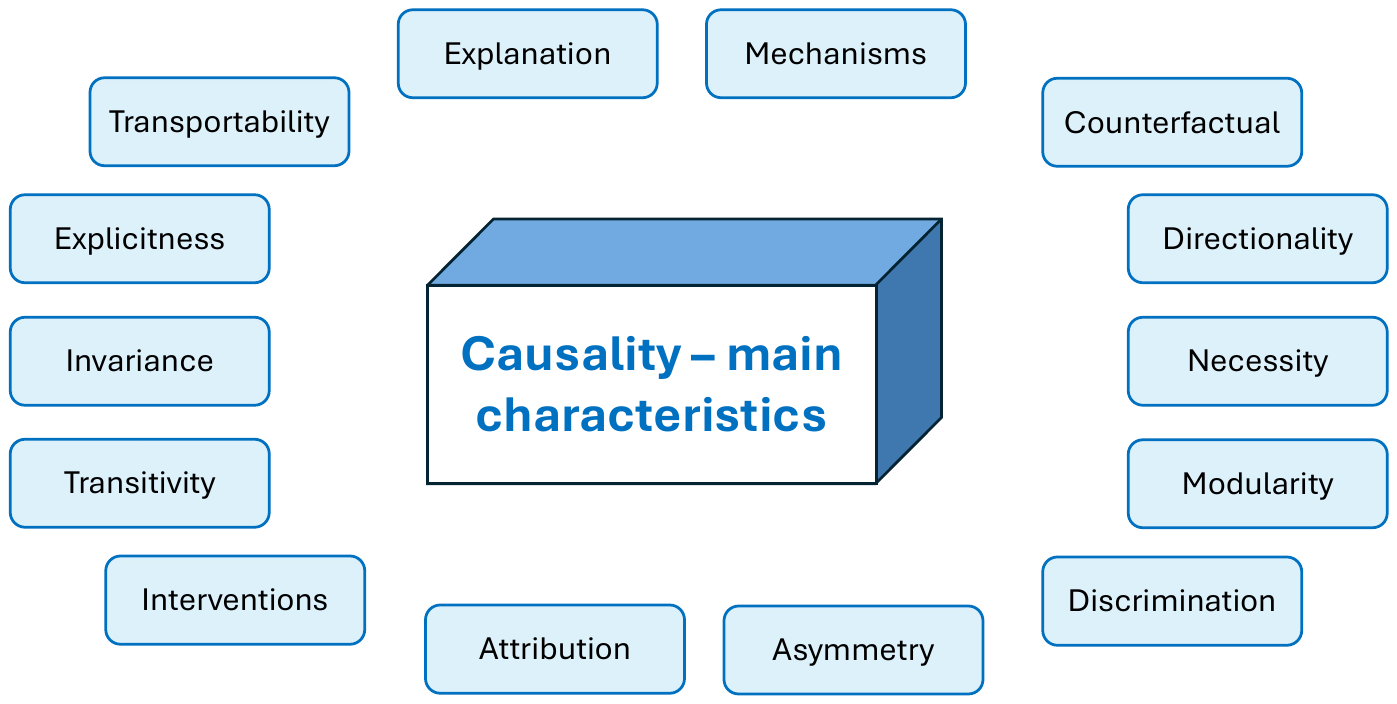}
	\end{center}
	\caption{Main characteristics, as recognized by the relevant literature, that makes Causality distinguishable of other AI domains.}
	\label{fig:CausalityMainCharacteristics}
\end{figure}

Causality approach is different from other approaches such as ML, statistical correlation and significance, and descriptive methods. Although some of its characteristics can be found in other approaches, their combination and richness are unique in the context of causality.
Some of the important characteristics of causality that distinguish it include Directionality \cite{hosseini2021predicting}, Necessity \cite{nadathur2020causal}, Manipulability \cite{harinen2018mutual,woodward2001causation}, Asymmetry \cite{white2006causal}, Transitivity \cite{eells1983probabilistic}, Invariance \cite{bica2021invariant},  Explicitness \cite{irwin1980effects}, Explanation \cite{bertossi2020causality}, Counterfactuals \cite{lagnado2013causal}, and Transportability \cite{schenker2004causal,ganguly2023review}
In addition causality possesses several unique characteristics that distinguish it from other relationship such as Mechanisms \cite{befani2012models}, Modularity \cite{cartwright2001modularity}, Interventions \cite{greiner2011causal}, Discrimination and Attribution \cite{olesen2010distinguishing,schenker2004causal} as can be seen in Figure~\ref{fig:CausalityMainCharacteristics}:.\\

The phrase ``correlation does not imply causation" is a well-known concept in research and statistics \cite{ksir2016correlation}. It highlights the important distinction between correlation and causation \cite{guyon2008practical}. One key factor that sets causality apart from correlation is directionality. Causal relationships involve a clear sense of direction, indicating that the cause precedes the effect \cite{hosseini2021predicting}.
In contrast, correlations for instance, lack inherent directionality and can go either way.
Another distinguishing characteristic are sufficiency and necessity \cite{PearlMackenzie18,Halpern16,nadathur2020causal}. Causality implies that a cause is either sufficient, necessary, or both for its effect to occur. Correlations, on the other hand, do not imply necessity or sufficiency and can exist without a cause-effect relationship.
Manipulability is a defining feature of causal relationships \cite{harinen2018mutual,woodward2001causation}. Causality allows for manipulation or intervention on the cause, resulting in observable changes in the effect. This characteristic provides researchers with the ability to test and examine causal relationships more directly.\\
Asymmetry is a fundamental distinction between causality and correlations. The causal relationship between two variables, such as `A' causing `B', is distinct and asymmetric from the reverse relationship of `B' causing `A' \cite{white2006causal}. In contrast, correlations exhibit symmetry, measuring the statistical relationship between variables without implying causation.
Transitivity, another property of causality, allows for chaining causal relationships. If `A' causes `B' and `B' causes `C', then `A' is considered an (indirect) cause of `C'. This concept of transitivity provides a deeper understanding of the interconnectedness of causal relationships \cite{eells1983probabilistic}. Correlations, on the other hand, do not inherently possess transitivity.
Invariance is a characteristic unique to causality. Causal mechanisms remain consistent and invariant under different interventions or contexts \cite{bica2021invariant,befani2012models}. Assuming the stability of the underlying causal structure, this property allows researchers to make reliable predictions and draw conclusions about causal relationships, whereas correlations can change based on specific circumstances.
Causal theories also emphasize explicitness by making transparent assumptions about underlying mechanisms. This explicitness goes beyond observed correlations and enables a deeper understanding of how and why causal relationships occur \cite{irwin1980effects}.\\
Explanation is an essential aspect of causality. Causality seeks to explain effects in terms of their underlying causes, rather than merely identifying patterns in data. By understanding the causal mechanisms at play, we can gain a more comprehensive understanding of the phenomena being studied \cite{bertossi2020causality}. Causality explanation characteristic increases the trustworthiness. This can be one of the reasons that causality is popular among decision makers.
Counterfactuals play a crucial role in causal reasoning. Causes support reasoning about hypothetical interventions using counterfactuals, allowing researchers to explore the effects of different scenarios \cite{lagnado2013causal,PearlMackenzie18,pearl2009causal}. In contrast, correlations do not provide the support for counterfactual reasoning.
Causality allows for transportability of knowledge. Valid causal inferences can extend beyond the specific conditions of observation due to structural or mechanistic assumptions. This property enables the application of causal knowledge to different contexts or populations \cite{Bareinboim_2013,ganguly2023review}.\\
Another characteristic that sets causality apart is the concept of modularity. Complex causal systems can be broken down into autonomous modules, each with defined inputs and outputs \cite{cartwright2001modularity,PetersJanzingSchoelkopf17}. By analyzing the compositionality of these modules, researchers can gain insights into the overall causality of the system. This modular approach provides a structured and systematic way of understanding complex causal relationships.
Interventions play a vital role in establishing causation. Through experiments and - under certain circumstances - natural experiments, researchers can isolate and manipulate proposed causal factors while controlling for other variables \cite{greiner2011causal}. This rigorous approach allows for the discovery of causal relationships through observations made under controlled conditions. Interventions enable researchers to go beyond mere associations and give a platform to falsify causal hypotheses.
Furthermore, causality provides a framework for attribution. It allows us to attribute responsibility, credit, or blame for the occurrence of certain effects. Unlike correlation alone, which merely identifies associations, causality enables us to assign causal responsibility and understand the consequences of specific causal factors \cite{schenker2004causal}.

\section{Causality Types, Relationships and Inference}
\begin{figure}
	\begin{center}
		\includegraphics
        [trim=3.5cm 8.0cm 12.0cm 12.0cm, scale=0.7]{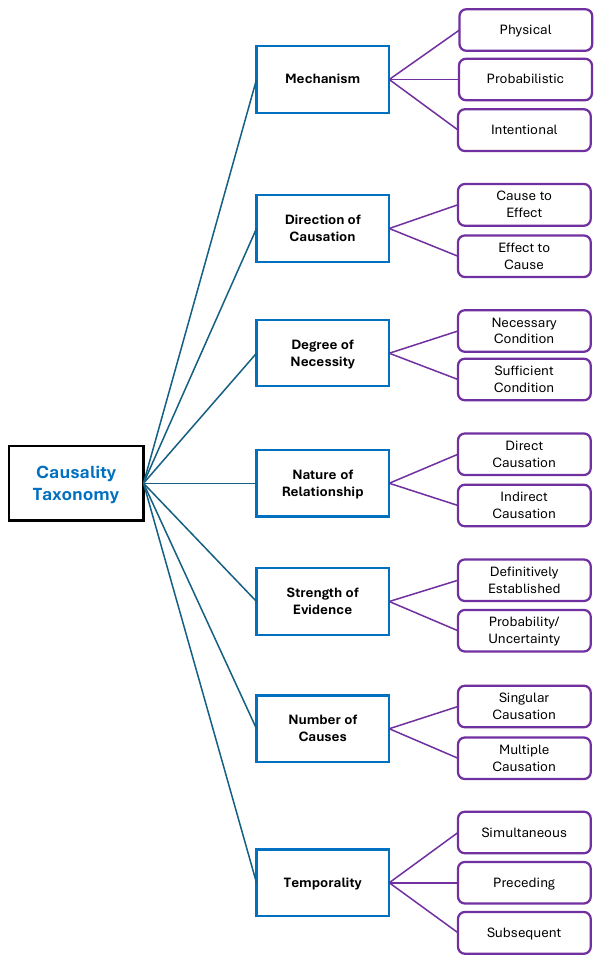}
	\end{center}
	\caption{Causality Taxonomy based on seven representative categories: mechanism, direction, necessity, relationship, evidence, causes, and temporality.}
	\label{fig:CausalityTaxonomy}
\end{figure}

Causality Inference (CI) refers to the process of identifying and understanding causal relationships between variables or events \cite{pearl2009causal}. 
Causal Discovery (CD) involves the analysis and construction of models that depict the inherent relationships within the data, while causal inference seeks to examine the potential effects resulting from altering a specific system \cite{yao2021survey}.
It involves determining whether one variable or event directly or indirectly influences the occurrence or outcome of another. 
We can find several approaches to causality as can be seen in Table \ref{tab:causal-inference-approaches}. 
\begin{table} 
\centering
\begin{tabular}{ |p{2.5cm}|p{2.27cm}|p{2cm}|p{2cm}|p{2cm}|p{2.1cm}|  }
\hline
\multicolumn{6}{|c|}{Causality Approaches} \\
 \hline
 \rowcolor{blue!30}
Approach& Conceptual Framework & Data Requirements & Assumptions & Estimation Methods & Interpretation\\
 \hline
  \hline
Causal Graphical Models (CGM) \cite{pearl1993bayesian}, Hidden Confounding (HC) \cite{ghassami2021causal}
& Graphical representation     & Joint distribution of variables & Acyclicity, no unmeasured confounders, ignorability (no unobserved confounders)
& Non-parametric identification and estimation                         & Direct interpretation based on graphical model structure                   \\
\hline
Potential Outcome (PO) Framework \cite{rubin2005causal} & Counterfactuals & Potential Outcomes             & Stable Unit Treatment Value Assumption (SUTVA), ignorability (no unobserved confounders)   & Matching, regression, weighting, etc.                   & Differences in potential outcomes under different treatment conditions     \\
\hline
Difference-in-Differences (DiD) \cite{o2016estimating}& Comparing changes over time  & Treatment and control groups   & Parallel trends between treatment and control      & Regression models with interaction terms, fixed effects & Average treatment effect based on comparison of changes over time             \\
\hline
Instrumental Variables (IV) \cite{angrist1996identification}    & Exploiting instrumental variables & Instrumental variables  & Relevance and validity of instrumental variables   & Two-stage least squares, instrumental variable regression & Causal effects based on association between instrumental variable and treatment \\
\hline
Structural Equation Modeling (SEM) \cite{fan2016applications,pearl1998graphs}& Modeling structural relationships & Data for estimating model parameters & Model-specific assumptions under causal identification               & Maximum likelihood estimation, other SEM-specific methods & Causal effects based on estimated parameters of the structural model \\ 
\hline
\end{tabular}
\caption{Comparison of causal inference approaches, highlighting the conceptual basis, requirements, assumptions, estimation methos, and interpretation capabilities.}
\label{tab:causal-inference-approaches}
\end{table} 
Causal Graphical Models (CGM) \cite{pearl1993bayesian} use graphical representation and do-calculus, allowing for direct interpretation of causal effects. The Potential Outcome Framework \cite{rubin2005causal} considers counterfactuals. 
Difference-in-Differences (DiD) \cite{o2016estimating}  compares changes over time between treatment and control groups. Instrumental Variables (IV)  \cite{angrist1996identification} exploits instrumental variables to estimate causal effects. Propensity Score Matching (PSM) \cite{caliendo2008some,li2013using} balances covariates and compares outcomes between matched individuals. Structural Equation Modeling (SEM) \cite{fan2016applications,pearl1998graphs} estimates causal effects based on the parameters of a structural model. Each approach has its own conceptual framework, data requirements, assumptions, estimation methods, and interpretation of causal effects.\\
Causality encompasses several main types that support understanding the relationships between causes and effects \cite{trampusch2016between,hood1979and}. These types shed light on the various ways in which these relationships can manifest. There are some popular types such as Direct or Strong, Indirect or Weak, Necessary, Sufficient, Multiple or Joint, Probabilistic, Common cause, Reverse or Spurious, and causal Homeostasis.\\
Direct/strong causality occurs when A directly causes B without any intermediary factors \cite{duan2012detection}. An example a pool cue (A) striking a billiard ball directly causes the ball to roll across the felt (B). 
Indirect/weak causality occurs when A causes B, but there are intermediary factors involved \cite{dufour2010short,vakorin2009confounding}. For instance, poverty indirectly causes poor health outcomes through factors like lack of access to healthcare and nutritious food.
Necessary causality signifies that A is necessary for B to occur, but it may not be sufficient on its own \cite{koutsoyiannis2022revisiting}. For example, lack of oxygen is necessary for death to occur, but other causes can also lead to death.
Sufficient causality indicates that A by itself is sufficient to cause B \cite{dul2016necessary}. For example, a high enough dose of poisoning may be sufficient to cause death without any other factors involved.
Multiple/joint causality arises when both A and B are required together to cause an effect \cite{petraitis1996inferring}. For example, A gene mutation (A) and environmental trigger (B) are necessary for some diseases to manifest.
Probabilistic causality suggests that A increases the probability or risk of B, but it does not guarantee the occurrence of B \cite{pearl1996structural}. For instance, obesity increases the risk of diabetes, but not everyone who is obese will necessarily develop diabetes.\\
Common cause causality occurs when a third variable, C, causes both A and B. For example, genetics C may cause both obesity A and diabetes B.
Reverse/spurious causality emphasizes that the assumed cause A may actually be caused by or correlate with the effect B, rather than vice versa \cite{tyagi1993exploratory}. It is important to correctly identify causes and effects to avoid misinterpretation.
Additionally, there is the concept of causal homeostasis. Causal homeostasis refers to the tendency of causal systems to resist change and maintain stability, even when external interventions or perturbations occur \cite{keil2006explanation,haggqvist2005kinds,weinberger2021}. This stability is achieved through feedback loops and networked causal mechanisms within systems that absorb, redistribute, or redirect incoming influences. As a result, interventions may have more limited effects than expected, as homeostatic regulatory processes work to return causal dynamics to their baseline state. Understanding causal homeostasis provides insights into why observational and interventional distributions, even when standard causal assumptions are violated.
It is important to recognize that causality is often probabilistic rather than definite, with various factors and complexities at play in cause-effect relationships. \\
Various types of causal relationships exist, each with distinct characteristics and implications. 
Confounding relationships can arise when third variables influence both the exposure and outcome. For example, stress, diet and exercise habits are interrelated - higher stress can reduce healthy behaviors but poor diet and exercise can also increase stress levels through physiological pathways. At the same time, all three influence health outcomes like heart disease risk. Attempting to isolate any direct causal effect, such as between stress and heart disease, is complicated by the bidirectional links between stress, diet, exercise and their collective impacts on health \cite{vakorin2009confounding}. Spurious relationships, on the other hand, are correlations that seem causal but are actually due to confounding factors \cite{gunter2012causal}. Mediating relationships occur when the effect of one variable on another is partially transmitted through a mediator \cite{imai2010general}, such as smoking cessation programs reducing lung cancer by reducing smoking. Moderating relationships occur when the strength or direction of a causal effect depends on a third variable, such as social support moderating the effect of stress on mental health \cite{wu2008understanding}. Lastly, bidirectional/recursive relationships involve two variables causally influencing each other in an ongoing feedback loop over time \cite{blalock2018causal}, for example, optimism and life success reinforcing each other. To accurately distinguish these different types of relationships from observational data alone, it is crucial to identify the causal graphical structure and apply appropriate causal inference techniques.\\
As shown in Figure~\ref{fig:CausalityTaxonomy},  we offer a taxonomy based on the following classes: Direction, Necessity, Relationship Type, Evidence Strength, Number of Causes, Temporal Sequence, Mechanism. 
Direction describes whether causes happen before, after, or simultaneously with their effects. Necessity distinguishes causal factors by their level of necessity or sufficiency to produce an effect. Relationship Type characterizes the link between cause and effect as direct, indirect, or spuriously associated. Evidence Strength categorizes the conclusiveness of empirical evidence supporting a causal claim. Number of Causes addresses whether one or multiple factors jointly contributed to bringing about an effect. Temporal Sequence considers the timing of causes and effects as simultaneous, cause preceding effect. Mechanism refers to the underlying processes or means through which causation is achieved, such as physical, probabilistic, and intentional.

\section{Causality Applications and Usages}
Causality plays a crucial role in various fields, impacting numerous applications. In medicine and healthcare, understanding disease etiology, evaluating treatment effectiveness, and precision medicine are important applications. Additionally, causality aids in clinical decision support, predictive risk modeling, and epidemiology \cite{alvarez2004causality}. \\
Transportability is a crucial aspect where causal models are needed to generalize across different contexts and populations \cite{hernan2011compound,prosperi2020causal}. This is particularly relevant in healthcare, where diverse groups may require tailored interventions and treatments. Heterogeneity is another key consideration, as causal understanding helps identify the treatments that work best for specific individuals. Precision medicine heavily relies on causal analysis to determine personalized approaches to healthcare \cite{kosorok2019precision}. Experimentation is an area where causal modeling excels, enabling the optimization of experimental design when fully randomized experiments are either unethical or impractical to conduct. Time series analysis benefits greatly from causal modeling as it allows for the examination of dynamic processes unfolding over time \cite{hlavavckova2007causality}. This is essential for accurate forecasting in various applications.\\
Moving to the realm of economics and finance, causality is essential for assessing the effects of policies and regulations, predicting the impact of economic decisions, and understanding consumer choice factors \cite{atanasov2016shock}. It also plays a role in forecasting economic trends \cite{moraffah2021causal}.
Mechanism design utilizes causal models to develop incentives and policies that shape desired outcomes. This application is commonly seen in the field of economics \cite{gow2016causal}. Fairness becomes an important consideration in algorithmic decision-making, and causal reasoning plays a significant role in identifying unfair biases and understanding their root causes \cite{kusner2017counterfactual}. The ability to answer what-if questions about interventions is critical for decision support \cite{pearl2019seven}. Causal models allow for the assessment of the potential impacts of different interventions, aiding in informed decision-making. 
In the field of education, causality helps in determining education interventions and reforms, implementing personalized learning approaches, and assessing teaching methods \cite{murnane2010methods}. It aids in understanding attrition factors and enables predictive analytics for educational outcomes. Public policy benefits from causality by evaluating the effectiveness of social and environmental programs, assessing the impact of laws and bills, and facilitating evidence-based decision making \cite{zajonc2012essays,yao2021survey}. \\
Causality is also significant in recommender systems \cite{wang2020causal,gao2022causal}, allowing for the explanation of recommendations, counterfactual reasoning for fairness, and estimating the effects of exposure to certain items. In fraud and anomaly detection \cite{vivek2022explainable}, causality helps in isolating causative factors for abnormalities rather than relying solely on correlations. It also contributes to predictive maintenance strategies. \\
In the field of robotics and control \cite{smith2020counterfactual,xu2023causal}, causality is crucial for causal world modeling, decision making under interventions, and simulating the downstream effects of actions. Sociology benefits from causality by modeling influence and diffusion networks, understanding shifts in group behavior, and exploring social determinants of various outcomes \cite{gangl2010causal}.\\
In marketing and advertising, causality aids in estimating campaign effectiveness, targeting high-potential customer segments, and optimizing acquisition and retention strategies \cite{varian2016causal}. In the domain of cybersecurity, causality plays a role in predicting vulnerabilities and risks, tracing the root cause of intrusions and attacks, and evaluating alternative security control measures \cite{abel2020applications}.
Explainability is a major advantage of causal models as they provide explanations in terms of cause-effect relationships rather than mere correlations. This enhances the transparency and interpretability of AI systems \cite{kuang2020causal}. \\
Counterfactual analysis, which involves generating realistic alternatives, is valuable for assessing the impacts of decisions that were not taken. This helps in understanding the potential outcomes of different choices.
Complex systems often require combinatorial models that involve causal reasoning across different formalisms. This interdisciplinary approach allows for a comprehensive understanding of intricate systems. In summary, causality plays a critical role in facilitating transfer learning, ethical and fair decision-making, policy optimization, and the development of transparent, explainable, and trustworthy AI systems \cite{rohmatillah2021causal}.
Causality is a fundamental concept that finds applications in various fields. Science relies on causality to establish cause-and-effect relationships between variables. \\
In philosophy, causality is a central concept in metaphysics and epistemology \cite{machamer2004activities}. The legal field employs causality to determine liability and responsibility for damages or injuries \cite{young2007causality}. In medicine, causality is crucial for understanding disease causes and developing treatments \cite{alvarez2004causality}. In the social sciences, causality is employed to understand social phenomena such as poverty, crime, and inequality \cite{marini1988causality}. \\
Business leaders utilize causality to understand the causes of success and failure, develop strategies for improving performance, and drive customer behavior. Economics employs causality to understand relationships between economic variables. Environmental science employs causality to understand the causes of environmental phenomena.
Overall, causality is a fundamental concept used across various fields. It helps understand relationships between variables, establish cause-and-effect connections, and develop interventions and strategies to address challenges in different domains.

\section{Preconditions and Architecture for Implementing Causality on Datasets}
Establishing causality from datasets is a complex task that requires careful consideration. In ML, defining the target variable and selecting relevant features are crucial steps in the implementation process. The choice of ML models plays a significant role, as they can be trained to improve the chances of obtaining the desired results. Nonetheless, when aiming to establish causality from datasets, it is essential to acknowledge and address several key preconditions. These preconditions, which help ensure the validity and reliability of causal inferences, involve considerations such as temporal order, confounding factors, and the presence of a plausible mechanism \cite{runge2019detecting}.\\
The database architecture also influences the causal inference process. Graph databases, due to their ability to represent Directed Acyclic Graphs (DAGs), provide a suitable platform for causal analysis. The DAG property of graph databases enables faster causal inference compared to popular relational databases. Additionally, graph databases can be effectively used to represent and analyze causality \cite{khoo2000extracting}.\\
The causal relationships can be effectively represented in a graph database by modeling them as directed edges or links between nodes, allowing for the explicit depiction of connections between objects. This approach enables the tracing of causal pathways and influences, empowering the analysis of connectivity and propagation within complex systems through graph queries and algorithms. By identifying interconnected causal networks and understanding how causal power propagates, it becomes possible to determine all entities indirectly impacted by a specific event or condition. In addition, ML techniques such as network embedding can be employed to infer unknown causal links by analyzing patterns of connectivity within a causal graphs. Graph databases also enable the simulation of what if scenarios, as they allow for the modification of causal link attributes or the selective addition or removal of nodes and edges.\\ 
This facilitates the exploration of how a causal network may behave under varying conditions or interventions. Moreover, the contextualization of causality is supported by assigning properties to nodes and edges in the graph, representing contextual factors that influence or modify causal relationships. This allows for the modeling of more nuanced and conditional forms of causality. Graph databases also provide a unified structure that facilitates the integration of diverse causal data types, such as experimental, observational, temporal, and spatial data, into a single queryable network, offering a comprehensive approach to studying causality. The Graph database architecture provides an optimal architecture that makes the causal inference process more efficient.\\
As mentioned above, there are inherent properties that are essential for casual inference. Firstly, temporal precedence is crucial, meaning that the potential cause must occur before the alleged effect, establishing a clear temporal order. Secondly, there needs to be a statistical association or correlation between the potential cause and effect variables in the data. Additionally, it is important to rule out confounding variables, alternative explanations, or common causes for the relationship. This can be achieved through statistical control. Furthermore, a plausible mechanism should be understood, based on scientific theory or knowledge, to explain how the cause could lead to the effect. It is essential to note that correlation does not imply causation \cite{subbaswamy2018counterfactual}.\\
The gold standard for establishing causality is experimental data, where the cause is intentionally manipulated to confirm its impact on the effect. Observational data, while valuable, is considered weaker. Another important factor is the presence of a dose-response relationship \cite{zhang2016causal}, wherein larger or smaller doses of the cause correspond to larger or smaller magnitudes of the response. Consistency across studies, contexts, and populations strengthens the evidence of causality, as it replicates the relationship.\\
Specificity is another precondition, indicating that a singular cause should lead to a specific effect and not a broad range of variables. Overly broad conclusions weaken the strength of the causal relationship \cite{fox1991practical}.
Establishing causality requires carefully designed studies, appropriate statistical techniques, and the elimination of threats to validity. Merely possessing large, correlated datasets is insufficient for determining causation. \\
Additional preconditions for establishing causality from datasets include considering sample size. Larger sample sizes are required to reliably detect even modest causal effects, control for confounding variables, and replicate findings \cite{king2017balance}. Insufficiently powered studies cannot rule out alternative explanations. Data quality is also critical, necessitating accurate, complete, and multi-variate data with minimal missing information. Measurement error, selection bias, and attrition can undermine causal conclusions.
Counterfactual analysis is vital, as it involves considering what would have occurred in the absence of the supposed cause. This is challenging without experimental manipulation of potential causes. Observational studies require stronger assumptions.
As we discussed before about the Graph database, DAGs can assist in formalizing assumptions about causal structure, identifying confounding variables, and determining if the data and research design can establish causality between specific variables.
Replication is pivotal, as causal conclusions gain strength when multiple independent studies using different datasets and methods yield consistent effects, minimizing the likelihood of chance findings \cite{sheffield2020replication}.
Theoretical consistency is another important factor, as causal hypotheses must align with scientific theory. Anomalous findings should be scrutinized carefully, as they may not reflect true causality \cite{robins2003uniform}.
Establishing causality from observational data requires assembling converging evidence from multiple angles and addressing threats to validity. It is a gradual and multifaceted process, rather than an absolute conclusion.

\section{Causality Evaluation and Metrics}
One of the most important ways to relate to the extracted results is to use set of criteria and evaluation metrics.
No single metric quantifies causal understanding; rather a combination provides a holistic evaluation \cite{zimmermann2021well}. The goal metrics focus on are also application dependent.
The previous section mentioned the essential datasets characteristics that enable casual inference. In this section, we consider the way to evaluate the extracted causality model. Before diving into the key aspects and metrics, there is a need to emphasise some important datasets characteristics that are support the model evaluation. The ideal dataset has to include both observational and  interventional/experimental data. In addition, causality evaluation has to be based on on multiple relevant datasets in order to test generalization.\\
In order to assess causal models metrics such as effect sizes, causal discovery accuracy, and counterfactual quality should be considered.
To establish a baseline for comparison, it is recommended to compare causal models to non-causal ML models. It is also valuable to evaluate them against causal inference methods like potential outcomes and matching. Another important factor is sample complexity. Testing how data requirements scale up for increasingly complex and unidentified causal problems helps understand the limitations and scalability of the models.
Assessing the robustness of results to violations of modeling assumptions is another critical aspect. Analyzing how sensitive causal inferences are to choices in the model, metric, hyperparameters, and training data through sensitivity analysis is necessary. Furthermore, evaluating the interpretability of models is essential to determine if they provide intelligible explanations for predictions and facilitate scientific insights.\\
Models should also be evaluated regarding how well they convey uncertainty about their causal conclusions. Benchmarking the transportability of models' causal conclusions to new domains and populations is crucial to understand their applicability beyond the training data. For dynamic or time series problems, it is important to test the models' ability to discern short-term versus long-term effects.
Using feedback is also recommended in the model evaluation process. For more complex problems with interactive and bidirectional causal relationships can provide insights into the models' performance in real-world scenarios. Thorough evaluation requires testing across multiple dimensions on standardized benchmarks using both simulated and real-world data. It is important to note that no single study can definitively validate a causal approach.\\
Several common metrics can be used to evaluate causal models and causal inference. Some of them are relevant to models such as ML while others are tailored to causality. These metrics include prediction error, intervention accuracy, effect estimate accuracy, confidence calibration, ROC AUC, structural Hamming distance, modularity score, causal sufficiency, transportability error, counterfactual sample quality, and counterfactual effect consistency. Each of these metrics offers a different perspective on the models' effectiveness in capturing causal relationships and estimating causal effects.\\
In addition to these evaluation methods, Hill's first criterion can be used to assess the strength of causal connections, specifically the association connection. Hill's criteria, primarily focused on health and biological perspectives but applicable to other causal associations, include Consistency, Plausibility, Temporality, Dose-response relationship, Experimental evidence, and Mechanistic evidence.\\
We already discussed the hierarchy of causal inference and its layers or rungs. By following this hierarchical approach, causality evaluation and metrics help quantify the strength of causal inference and provide valuable insights into the nature of causality.
While there is no universally agreed-upon hierarchy of causal inference methods, some common frameworks discuss causality approaches in a hierarchy or order of strength \cite{duckworth2010establishing}. The hierarchy generally progresses from weaker association-based methods using observational data to consistency across studies, plausibility evaluations, establishment of proper temporality and dose-response relationships, and culminates in the strongest evidence from experimentation and mechanistic understanding of the processes involved \cite{cook2002experimental}. These later approaches help rule out alternate interpretations and provide confirmatory evidence.

\section{Causality Trustworthiness}
Ensuring a model can be trusted is vital for responsible deployment, responsible use, and continued uptake and reliance on the model. It helps address various legal, ethical, and practical concerns regarding AI safety and transparency \cite{ganguly2023review}. The need for trustworthiness intensifies when the model is extracted without human intervention such as in the case of causality models. As already mentioned above, the causality models by their nature holds the trustworthiness characteristics that can be affirmed by human logic and common sense. The trustworthiness of causality also relates to its inherent ability to explained to its users and will be discussed in the explainablity section in detail.\\
When evaluating the trustworthiness of causal claims and models, it is important to consider several key aspects. These aspects provide insights into the reliability and validity of the reported causal conclusions \cite{liu2023towards}.
Firstly, data quality plays a critical role, as potential biases, omissions, or errors in the observational data can undermine the accuracy of causal inferences. Additionally, the identifiability of the causal structure and assumptions from the given data is crucial, as the lack of identifiability can impact the validity of the conclusions.
Confounding is another vital aspect to address, ensuring that all common causes of the treatment and outcome variables are adequately measured and accounted for to avoid biased estimates. The appropriateness of the assumed causal graph structure and model form, known as model specification, should also be carefully examined, allowing flexibility to validate assumptions.\\
Accurate parameter estimation is essential to determine the precision and accuracy of the estimated causal effects from the data, while being cautious of the potential for overfitting. Transportability is another consideration, ensuring that the estimates can be generalized beyond the observed context and are not contingent on unmeasured factors. \\
Sensitivity analysis provides an understanding of the robustness of the conclusions to violations of assumptions through what-if analysis. Predictive performance evaluation focuses on how well the estimated effects can predict the outcomes of planned interventions.
Transparency is a crucial aspect of establishing trust, requiring clear specification of assumptions, limitations, and methodological details to enable scrutiny. Lastly, evaluation should go beyond accuracy metrics and consider the suitability of the conclusions for intended uses, incorporating valuable stakeholder feedback.

\section{Causality and Explainable AI (XAI)}
Explainable AI (XAI) refers to the research and methods aimed at providing transparency and understandability to the decision-making process of artificial intelligence models \cite{arrieta2020explainable}. It addresses the challenge of interpreting the outcomes and inner workings of complex AI systems, which are often perceived as black boxes. Causality, on the other hand, can serve as a surrogate model for these black box AI models, enhancing their level of explainability. By combining causality and XAI, a more comprehensive understanding of AI decision-making can be achieved.
Causality and Explainable AI (XAI) are two interconnected fields that have garnered significant attention in recent years. Causality serves as a surrogate model for AI black box models, augmenting their level of explainability. By combining causality and XAI, a more comprehensive understanding of AI decision-making can be achieved \cite{janzing2020feature}.\\
One aspect where causality and XAI intersect is in providing causal explanations. XAI endeavors to shed light on the decision-making process of AI models, while causality helps identify the causal relationships between variables that contribute to these decisions \cite{naser2021engineer}. By merging these disciplines, it becomes possible to generate explanations that not only describe the contributing factors but also elucidate the causal connections between them.\\
Another area of synergy lies in causal attribution. Causal attribution involves assigning credit to the factors that influence a decision or outcome \cite{janzing2020feature}. XAI can pinpoint the most significant contributors to an AI model's decision, while causality can determine the causal relationships between these factors. Integrating causal attribution and XAI yields a more comprehensive understanding of how AI models arrive at their decisions.\\
Causal visualization is yet another domain where causality and XAI converge. Causal visualization entails creating visual representations of the causal relationships between variables \cite{holzinger2021explainable}. XAI can generate visual explanations of AI decision-making processes, while causal visualization captures the causal links between the contributing factors. By combining these approaches, more informative and intuitive explanations of AI decision-making can be crafted.
Moreover, the combination of counterfactual explanations and XAI offers valuable insights into AI models' decision-making processes \cite{chou2022counterfactuals}. Counterfactual explanations explore hypothetical scenarios by describing what would have transpired had a specific action or event not occurred. XAI can generate such counterfactual explanations, while causality helps identify the causal relationships that contribute to these decisions. This fusion allows for a more comprehensive understanding of AI models' decision-making.\\
Lastly, explainable RL benefits from the integration of XAI and causality \cite{madumal2020explainable}. RL trains AI agents to make decisions based on rewards or penalties. XAI can explain the decisions made by these agents, while causality uncovers the causal relationships between the agents' actions and the rewards or penalties received. By combining XAI and causality, a more comprehensive explanation of the decision-making process of RL agents can be provided.

\section{Causality and Machine Learning (ML), AI, Genetic Algorithms (GA), and Generative AI (GAI) Approaches}

AI and ML approaches that have been used for solving problems and prediction have their own benefits and drawbacks in comparison to causality and causal inference in particular. This section will shed a light to the differences between the approaches and the interaction as well as combination between them.\\
According to Bishop et al. \cite{bishop2021artificial} the integration between AI and causality has its limitations, and simply relying on causal reasoning may not be sufficient to address those limitations. However, a review of the role of causality in developing trustworthy AI systems highlights the potential benefits of incorporating causality \cite{ganguly2023review}. Causal AI not only has the ability to predict but also offers the capacity to answer questions that traditional ML models cannot. Unlike predictive ML models, Causal ML explicitly accounts for confounders by modeling both the treatment of interest and its impact on the outcome \cite{ganguly2023review}. This allows Causal ML to isolate the causal impact of treatment on the outcome, going beyond mere correlation.\\
There are several key differences between causality-based approaches and traditional ML approaches. Firstly, while causal methods can often utilize observational data, ML typically requires experimental data to establish reliable cause-effect relationships. Secondly, ML focuses on prediction, while causality aims to understand the underlying generative processes and reason about new interventions. Causal graphs make explicit independence assumptions between variables, whereas ML primarily fits functional forms \cite{ganguly2023review}. Moreover, causal graphs and counterfactual analysis provide human-interpretable explanations of model predictions in terms of causes and effects, whereas ML models are often viewed as black boxes without clear causal interpretations.\\
Causality also offers advantages in terms of transportability, as causal inferences may generalize and propagate interventions beyond the training setting, whereas ML models typically do not. Causal structures are generally robust to interventions and conditional distributions, whereas ML models may fail under distribution shift. Additionally, causality helps address issues of fairness, bias, and discrimination by focusing on causal effects rather than associative effects \cite{ganguly2023review}.\\
As mentioned above, in terms of trustworthiness, causal models provide human-interpretable explanations through graphical structures, while ML models are often opaque. Causal assumptions are explicitly stated in causal models, while ML assumptions are implicit in the algorithm and data. Causal effects have the potential to generalize beyond training examples, while ML performance relies heavily on identically distributed test data. Causal graphs encapsulate invariance, while ML predictions can change unpredictably with small input changes. Causal methods can quantify and mitigate unfair biases, whereas ML models risk amplifying biases present in the training data. Causal estimates aim to remain valid under interventions, while ML models often do not transport beyond their training conditions. Lastly, causal conclusions account for violations of assumptions, while ML performance is highly sensitive to violations of Independently and Identically Distributed (IID) assumptions \cite{ganguly2023review}.\\
The Generative AI (GAI) models holds nowadays, an important roles in many fields. Although GAI is a subfield of AI, when comparing causality with GAI the task intensifies. GAI systems and humans exhibit key differences in their understanding of causality. Generative models like Generative Adversarial Networks (GANs) learn patterns and generate new data without a comprehensive causal model of the world, while humans combine data-driven learning with intuitive physical reasoning.
Generative models, such as GANs, excel at learning patterns and generating new data, but they lack a comprehensive causal model of how the world functions. In contrast, humans combine data-driven learning with intuitive physical reasoning.\\
AI causality primarily relies on statistics and correlations derived from large datasets, whereas humans infer causality through experimentation, reasoning about mechanisms, and contemplating counterfactuals, such as what-if scenarios.
GAI lacks subjective experience and commonsense knowledge regarding how causal factors like psychology, culture, and societal influences shape human behaviors and decisions.
AI focuses on identifying statistical causes to generate realistic outputs without a profound understanding of how physical, mental, and social factors interact in complex real-world phenomena continuously refine their understanding of causality through lifelong open-ended learning from diverse sources. Generative models, on the other hand, remain confined to their initial narrow training data and predefined objectives.
Human causal attributions involve reasoning about agency, responsibility, and intentions underlying events. In contrast, GAI operates solely on correlations without possessing higher-order faculties for such considerations.
Another type of algorithms, that relates to ML approach are Genetic Algorithms (GA). GA are optimization techniques inspired by biological evolution, aiming to replicate the natural selection process \cite{stewart2004genetic}. Rather than determining causal relationships or assigning causality, they focus on exploring complex search spaces and finding optimal solutions. By applying selection, crossover, and mutation principles to populations of potential solutions, GA iterate through generations and select the fittest candidates to drive the search. While crossover and mutation introduce random changes, mimicking natural genetic variations, genetic algorithms cannot directly infer causal relationships. In ML tasks, such as predicting relationships in data, GA identifies correlations that optimize the fitness function but do not establish causation. Additional analysis of GA results may allow for inferences about potential causal relationships, but determining causality requires controlled laboratory experiments or observational studies. The value of GA solutions lies in their fitness, not necessarily in accurately modeling the true causal structure that generated the data, if causality even applies. GA prioritize optimizing what works rather than explaining why it works causally.

\section{Causality and Bigdata}
Bigdata datasets have been rapidly growing in recent decades due to several technological trends that have increased storage volume, enabled efficient handling by powerful processors, and facilitated the use of advanced query languages and ML models. The bigdata trend encourages institutions and organizations to collect as much data as possible, creating a snowball effect where the more data they collect, the larger data platforms they require. \\
Causality presents significant challenges in the context of bigdata, as causal inference must operate on a large volume of data that is beyond human handling capabilities. Additionally, it needs to accommodate the characteristics of big data known as the 5V's: velocity, volume, value, variety, and veracity \cite{demchenko2014defining}. This further intensifies the challenge of developing causality models, as it necessitates not only handling vast amounts of data but also dealing with its variety and constructing a causal model within a limited timeframe.
In addition to the above digital interventions, such as A/B testing of user interface modifications and personalized recommendations at scale, offer opportunities to conduct large-scale experiments and uncover cause-effect relationships within real user behavior.\\
The understanding of causality in the realm of bigdata analysis involves several key aspects \cite{guo2020survey}.
Observational data, which comprises large passively collected datasets, has facilitated the exploration of complex causal structures that were previously too intricate or costly to analyze. However, the quality of inferences heavily relies on the ability to effectively control for confounding factors.
New statistical techniques \cite{mannering2020big}, such as graphical models and structure learning algorithms, are being scaled up using methods like subsampling, approximation, and distributed computing. These advancements aim to handle the complexities associated with thousands of variables in causal analysis.
Passive observational studies utilizing extensive patient datasets have emerged as viable alternatives to randomized experiments when identifying treatment effects, thanks to the availability of adequate sample sizes. However, the concern of selection bias still persists.\\
The combination of multiple datasets, encompassing diverse sources like electronic health records, social media, and mobile data, can provide complementary insights. On the other hand, addressing challenges related to data alignment and integration is crucial.
Validating predictions and quantifying uncertainties at scale have become feasible due to the ability to re-identify individuals and capture finer-grained attributes. However, this increased capability raises privacy risks that must be carefully managed.
Causal discovery in the context of bigdata poses challenges, particularly in scaling techniques for complete identification of causal structures. Additionally, data integration and the selection of relevant variables remain ongoing obstacles in the pursuit of comprehensive causal analysis.

\section{Causality and Reinforcement Learning (RL)}
Reinforcement Learning (RL) is a branch of ML where an agent learns to make decisions by interacting with an environment. It involves the agent taking actions, receiving feedback in the form of rewards or penalties, and using this feedback to improve its decision-making over time \cite{arulkumaran2017deep}. RL allows automated agents to learn optimal behavior directly from experience to solve complex real-world challenges through trial-and-error interaction. Its usage is increasingly prevalent nowadays due to emerging of new algorithms and usages such as in Large Language Models (LLM) application as ChatGPT. \\
Causality and RL are connected through their shared goal of decision making through modeling systems and both attempt to address challenges by leveraging observational data. Causality can enhance RL if incorporated appropriately.
Causality RL is an interdisciplinary research field that merges concepts from causal inference and RL to investigate the causal effects of actions in decision-making processes \cite{guo2020survey}. Its objective is twofold: to learn a policy that maximizes cumulative rewards and to identify the causal relationships between actions and environmental changes.\\
In typical RL scenarios, agents learn to make decisions by interacting with the environment and receiving rewards or penalties \cite{sontakke2021causal}. Nonetheless, in real-world situations, the causal relationships between actions and environmental changes are often unclear, necessitating the agent's ability to reason about causal effects for informed decision-making.
Causality RL addresses this challenge by integrating causal inference methods into the RL framework \cite{grimbly2021causal}. Agents use causal models to represent the causal relationships between actions and environmental changes, enabling them to infer the causal effects of their actions. This information guides the learning process, ensuring that the agent's policy maximizes expected rewards while aligning with observed causal data.\\
Key concepts in Causality RL include causal graphs, which depict the relationships between variables in the environment, and causal inference, which employs statistical methods and data to infer causal relationships \cite{lu2022efficient}. Counterfactuals play a role in reasoning about the effects of alternative actions, and causal loops represent situations where the agent's actions impact the environment, which, in turn, affects the agent's rewards \cite{he2022causal}.
Causality RL offers several benefits, including improved decision-making by considering causal effects, better handling of confounding variables through inferred causal relationships, and increased transparency through explicit causal models.
To date, challenges persist within Causality RL \cite{guo2020survey}. The complexity of causal models can hinder interpretation and learning of causal relationships. Identifying causal relationships may be difficult when they are not readily apparent, and striking a balance between exploration and exploitation poses a challenge, particularly with limited causal information.

\section{Integration between Causality, Machine and Deep Learning, and Genetic Algorithms}
In the previous sections, the difference between causality and ML approaches was mentioned. This section will discuss the interaction and connections between the two. The section relates to Deep Learning (DL) as subfield of ML, unless mentioned otherwise.
Causality and ML are intertwined in several ways, leveraging ML techniques to uncover causal relationships between variables and make predictions based on these relationships. The relationship between causality and ML manifests in various aspects. 
Causal inference plays a crucial role, utilizing ML to deduce causal relationships by analyzing patterns in data \cite{scholkopf2022causality}. There are approaches of harnessing DL to transform distributions into a representation space such that they are indistinguishable \cite{ramachandra2018deep}.
Causal graphs provide a visual representation of causal relationships between variables. ML algorithms facilitate learning the structure of causal graphs from data by identifying patterns of causality among variables \cite{guo2020survey,kalainathan2020causal}.\\
ML algorithms enable the fitting of causal models to data. These models allow predictions about intervention effects on outcomes. Estimating counterfactuals, hypothetical outcomes if a specific treatment were assigned, is achievable through ML algorithms. By utilizing data from control groups, ML algorithms can estimate the outcome of a treatment for an individual \cite{prosperi2020causal}. Controlling for confounding variables, which can affect both the outcome and treatment, is achievable through ML algorithms. Including these variables as covariates in a regression model allows for confounding adjustment. 
Personalized medicine benefits from ML algorithms that tailor treatments to individuals based on their unique characteristics and causal relationships between treatment and outcome \cite{sanchez2022causal}. Genetic profiles and medical histories can be employed to predict the treatment's impact on a specific individual.
Causal discovery utilizes ML algorithms to unveil causal relationships between variables by analyzing data patterns \cite{arora2019bayesian}. For instance, genomics and clinical data can be used to identify causal links between genetic variants and diseases.\\
Additional commonly used approach in ML is ensemble methods. Ensemble methods involve combining multiple models or predictions to improve overall performance and make more accurate predictions \cite{dong2020survey}. This can be done through techniques such as bagging, boosting, or stacking. Ensemble methods have been successful in various domains and can be applied to both classification and regression tasks. One of the key ideas for using Ensmeble in the context of causality is to reduce uncertainty in causal inferences by averaging plausible estimates from diverse models encoding alternative assumptions.\\
Following by are several approaches for building causal Ensembles. Structural Causal Model (SCM) ensemble involves training multiple SCMs using different variable orderings and exclusion restrictions, and then taking a consensus from the ensemble \cite{li2020new}. Another approach is the Potential Outcomes ensemble, where individual treatment effect estimates are averaged from models that incorporate different sets of control variables \cite{austin2012using}.
The Counterfactual Gaussian Process Ensemble   utilizes Gaussian Process (GP) regression \cite{liu2019adaptive,mishler2021fade} models to establish relationships between treatments and outcomes, and the ensemble estimates the average causal effect by marginalizing over these models. Additional usage of ensemble that yields more accurate results can be found in decision causality trees and forests \cite{younas2022optimal} as well as causal rules \cite{lee2020causal}.
In the Heterogeneous Effect Modeling Ensemble, separate treatment-stratified models are built to estimate effect modifiers, and the subgroup estimates are then combined \cite{bargagli2020causal}.\\
Encoder-decoder is a common used architecture in DL. Encoder-decoder models have effectively been used for causality reasoning from natural language \cite{nayak2022generative}. Cause-and-Effect Pair Mining (CEPM) \cite{hassanzadeh2019answering} is an unsupervised model that learns relationships without labels by generating candidate pairs with an encoder-decoder and scoring them. Causal Inference over Natural Language (COIN) \cite{feder2022causal} also uses an encoder-decoder to produce implications and explanations, representing causality as queryable graphs. Some models not only propose claims but validate them via an encoder-decoder rationale generation component. Additionally, counterfactual text can be created with encoder-decoders by suggesting alterations to a potential cause or effect \cite{sampat2022learning}. These techniques represent causality and help justify relationships, revealing how encoder-decoders are well-suited for understanding causality from language.\\
ML algorithms enhance the interpretability of causal models by shedding light on the causal relationships between variables \cite{moraffah2020causal}. Identifying the most impactful causal relationships can be accomplished by analyzing the magnitude of their effects.
Time-series analysis benefits from ML algorithms to study causal relationships between variables that change over time \cite{huang2020detecting}. Economic data, for example, can be employed to predict the effect of policy interventions using ML algorithms.
ML algorithms facilitate causal inference even with incomplete or missing data. By leveraging observed data patterns, missing data can be imputed using ML algorithms.
The above mentioned approaches and examples exemplify how ML and DL can be harnessed to explore causality within data. Integrating ML with causal inference allows for unveiling causal relationships between variables, making predictions regarding intervention effects, and applying these insights in various domains such as healthcare, finance, and economics.
In previous sections, we mentioned the gap between GA and causal inference approaches. While there are approaches that tries to connect between the two \cite{li2019causality,blecic2007decision,de2000approximating}. 

\section{Causality and Generative AI (GAI)}
Causality plays a crucial role in explaining past results, but it can also support generated data. The integration between causality and emerging GAI plays an important role in research and applications. \\
Various GAI approaches incorporate causal reasoning, such as Causal Bayesian Networks \cite{deleu2022bayesian}, SCM \cite{pawlowski2020deep}, Counterfactual Models, Causal Trees \cite{darondeau1989causal}, and CausalGANs \cite{kocaoglu2017causalgan}. Furthermore, there are ongoing research areas related to causality, including Causal Discovery, Interventional Models \cite{shpitser2006identification}, Counterfactual Reasoning, Transportability, Causal Disentanglement \cite{yang2021causalvae}, Longitudinal and Time Series Data \cite{wunsch2010we}, and combining Causality and DL \cite{berrevoets2023causal}.
There is still much work to be done in scaling up causal learning methods to handle large, complex real-world datasets while maintaining properties like robustness, modularity, and interpretability. 
Several commonly used causal generative models include SCMs \cite{pawlowski2020deep}, Causal Bayesian Networks (CBNs) \cite{heckerman2013bayesian}, Potential Outcome Models/Counterfactuals \cite{lei2021conformal}, Markov Decision Processes (MDPs), Causal Additive Models (CAMs) \cite{buhlmann2014cam}, Generative Adversarial Networks with Backdoor Adjustment (GAN-BA) \cite{balazadeh2022partial}, Latent Causal Confounder Models \cite{louizos2017causal}, Causal Inference Neural Networks (CINN) \cite{luo2020causal}, and Self-Supervised Causal Discovery (SSCD) \cite{sontakke2021causal}.\\
These models formalize different causal assumptions to simulate interventions, estimate counterfactuals, and remove confounding bias to infer direct and indirect causal effects. Incorporating causal models into GAI can lead to advancements such as causal language models, structural causal Variational AutoEncoders (VAEs) \cite{kim2021counterfactual}, causal infilling, Conditional Generative Adversarial Network (cGAN) with causal supervision \cite{terziyan2023causality}, contrastive generation, causal video prediction, and counterfactual generations. These examples demonstrate how explicit causal reasoning brings interpretability, interventional reasoning, and robustness benefits compared to traditional associative generative models.

\section{Causality and Fuzzy Logic}
Causality and fuzzy logic are two interconnected concepts that have been extensively explored and integrated in various research studies and practical applications \cite{wang2014fuzzy}. One such concept is fuzzy causality, which combines causality and fuzzy logic to reason about causal relationships in situations where the precise knowledge of such relationships is lacking \cite{kim1998fuzzy}. Fuzzy causality provides a means to represent incomplete or uncertain information regarding the causal connections between variables, and it finds utility in decision-making processes and predictive modeling.\\
Another area of investigation is causal fuzzy systems, which merge causal inference with fuzzy logic to analyze intricate systems characterized by uncertain or imprecise causal relationships \cite{miao2000causal}. By employing fuzzy sets and fuzzy logic, causal fuzzy systems enable the representation of the causal associations between variables. They find application in diverse domains such as control systems, robotics, and medical decision-making.\\
Fuzzy causal models, on the other hand, are statistical models that utilize fuzzy logic to represent causal relationships between variables \cite{zhang2001fuzzy,chen2014quantitative}. These models prove valuable when dealing with situations where the causal connections are uncertain or imprecise, allowing the inference of causal relationships from observational data.
The research field of fuzzy decision-making combines fuzzy logic and decision-making to investigate the process of decision-making in scenarios with incomplete or uncertain information \cite{lin2008causal}. Fuzzy decision-making techniques find applications in various domains, including medical, financial, and environmental decision-making, where the imprecision or uncertainty of information plays a significant role.
Lastly, fuzzy cognitive maps are cognitive models that employ fuzzy logic to represent the causal relationships between variables within a decision-making process \cite{kosko1986fuzzy,miao2000causal}. These maps offer insights into decision-making under incomplete or uncertain information and aid in the design of decision-support systems capable of handling imprecise or uncertain data.

\section{Conclusions and Future Directions}
AI and ML today rely primarily on statistical correlations in data to infer causation, whereas humans understand causation through intuitive theories about how the world works based on mechanisms, context, and common sense knowledge. Humans also view causation dynamically and consider uncertainties like hidden or confounding variables. Most current AI/ML approaches take a limited, static view of causation directly from observational data alone. We estimate one of the future challenges is to close this gap by endowing AI with capacities that mimic human causal reasoning abilities more closely. Giving systems mechanisms for broader domain and common sense knowledge, causal discovery algorithms accounting for correlation vs causation, representing and reasoning about uncertainties and unobserved variables, and continually refining beliefs based on new evidence could help align how causality is understood by AI systems with the more robust and flexible understanding of causality humans naturally possess. \\
In the realm of causal inference, various research areas can contribute to narrowing the gap between AI/ML causality and human causality. One such area is causal representation learning, which focuses on developing methods to learn embeddings or representations of variables that encode causal relationships and mechanisms. By capturing semantic similarity related to direct or indirect causality, these models can assist in tasks such as causal inference and structure learning. However, real-world data's lack of perfect causal labels presents a challenge in this domain. 
Another important research avenue is causal time series modeling, which extends existing causal graph models to incorporate temporal dependencies and feedback loops in time series data. This extension allows for the modeling of dynamic changes over time, enabling unified frameworks for tasks like time series forecasting, anomaly detection, and counterfactual analysis. Addressing non-stationarities within the data presents a significant challenge in this field.
Causal reinforcement learning is another promising area, involving the integration of causal reasoning into reinforcement learning. By incorporating causality, agents can better understand how their actions propagate consequences in the environment. This integration has the potential to enhance offline evaluation, generalization, and safe exploration. However, developing scalable causal models to handle complex and dynamic environments remains a challenge. \\
Additionally, causal discovery at scale is a critical area of research. The goal is to develop scalable algorithms capable of handling large-scale datasets with hundreds or thousands of variables. Tech niques such as two-stage learning approaches, variable clustering before structure search, and random projection or sampling can enable causal analysis of complex real-world systems. Thorough validation using synthetic and real-world benchmarks is essential in this context.
By pursuing these research directions, we can strive to bridge the gap between AI/ML causality and human causality, fostering a deeper understanding of causal relationships and paving the way for more reliable and interpretable AI systems.

\section*{Acknowledgements}
We would like to thank Aleksander Molak for his feedback and remarks to the paper.

\bibliographystyle{IEEEtran}
\bibliography{ref.bib}

\end{document}